\documentclass[conference]{IEEEtran}

\usepackage[cmex10]{amsmath}
\usepackage{amsthm}
\usepackage{amssymb}
\usepackage{mathrsfs}
\usepackage{graphicx}
\usepackage{float}
\usepackage{array}
\usepackage{epstopdf}
\usepackage{multirow}

\usepackage{blkarray}
\usepackage{mathtools}
\usepackage{times}
\usepackage{subcaption}
\usepackage[T1]{fontenc}

\newcommand{\argmin}{\arg\!\min}

\ifCLASSINFOpdf
 
\else
 
\fi

\begin{document}

\title{Screen Content Image Segmentation Using Sparse-Smooth Decomposition}

\author{\IEEEauthorblockN{Shervin Minaee}
\IEEEauthorblockA{ECE Department\\
NYU  School of Engineering\\
shervin.minaee@nyu.edu}
\and
\IEEEauthorblockN{Amirali Abdolrashidi}
\IEEEauthorblockA{ECE Department\\
NYU School of Engineering\\
abdolrashidi@nyu.edu}
\and
\IEEEauthorblockN{Yao Wang}
\IEEEauthorblockA{ECE Department\\
NYU School of Engineering\\
yaowang@nyu.edu}}

\maketitle

\begin{abstract}
Sparse decomposition has been extensively used for different applications including signal compression and denoising and document analysis. In this paper, sparse decomposition is used for image segmentation. The proposed algorithm separates the background and foreground using a sparse-smooth decomposition technique such that the smooth and sparse components correspond to the background and foreground respectively. This algorithm is tested on several test images from HEVC test sequences and is shown to have superior performance over other methods, such as the hierarchical k-means clustering in DjVu. This segmentation algorithm can also be used for text extraction, video compression and medical image segmentation.
\end{abstract}

\IEEEpeerreviewmaketitle

\section{Introduction}
Image segmentation aims to decompose an image into multiple parts, making its processing easier and more meaningful.
One special case of this segmentation is background/foreground separation which tries to decompose an image into two components, background and foreground. This background/foreground separation can be used in various applications including: text extraction from document images \cite{text}, separate coding of background and foreground for compression of mixed content images \cite{MRC}, medical image segmentation and classification \cite{chromosome}, and line extraction for palmprint recognition \cite{palmprint1}.

The segmentation algorithm in this paper is mainly designed for screen content images which refer to images generated by the display screens of electronic devices. These images have some fundamental differences with photographic images and the usual image compression algorithms such as JPEG2000 \cite{jpeg} and HEVC intra-frame coding \cite{HEVC} may not result in a good compression rate for this kind of images. 
This is due to the existence of many sharp edges in this type of images. As a result, the energy cannot be compacted in a few coefficients and will be distributed in most of the transform domain coefficients, resulting in a high bit-rate.
In these cases, segmenting the image into two layers and coding them separately could be more efficient.
The idea of segmenting an image for better compression was proposed in DjVu algorithm for scanned document compression \cite{djvu} and the mixed raster content representation \cite{MRC}.

There are a few existing works for segmenting the image into background and foreground and coding them separately. 
Some of them use color clustering to perform the segmentation. Among them, DjVu \cite{djvu} is a popular document format which uses hierarchical k-means clustering for segmentation. It applies the k-means clustering algorithm with k=2 on blocks in multi-resolution. 
Some of the other works use morphological operations to separate the background layer from the foreground. In \cite{check}, Huang proposed one such algorithm for check image compression where he first segments the check image into different layers using the morphological operation and then code them separately.
Some other algorithms use simple color counting with some post-processing to perform background/foreground segmentation. Shape primitive extraction and coding (SPEC) \cite{SPEC} is an example of such algorithms. In SPEC a two-step segmentation algorithm is proposed to separate background from foreground. The first step uses the number of colors for segmentation and the second step will refine the results by extracting shape primitives (horizontal lines, vertical lines or rectangles with the same color).
Most of these approaches have difficulty for some of the screen content images where the foreground is overlaid over a smoothly varying background that has a large color range that overlaps with the color of the foreground and they usually tend to segment some part of the background as foreground.

In this work, a sparse decomposition framework is proposed for background/foreground segmentation that it is able to perform a correct segmentation even if the background and foreground colors overlap.
Sparse representation has been used for various applications in recent years, including face recognition \cite{wright}, super-resolution \cite{yang}, matrix decomposition \cite{rahmani2}, image restoration \cite{mairal}, detection and estimation \cite{jons1}, sparse coding \cite{jons2} and image classification \cite{mousavi1}. 
Despite the huge application of sparse representation, to the best of our knowledge, it has not been used for image segmentation so far.
The proposed algorithm in this paper is designed based on the intuition that in most of screen content images, background contains the smoothly varying part of the image, while foreground contains text, line and graphics creating sharp discontinuity.
Therefore background can be approximately represented by a smooth model, whereas the foreground cannot be modeled with this smooth representation.
Then the main effort is to find a suitable low-order smooth representation for background and classifying each pixel in the image into either background or foreground part. Since the foreground layer is expected to contain the high frequency part of the image, this algorithm can also be used for extraction of iris patterns from iris images \cite{iris1}, and also minutiae from fingerprint \cite{fing1}.

The rest of this paper is structured as follows. Sparse-smooth decomposition formulation is explained in Section II. Section III presents the overall segmentation algorithms. The experimental results and comparisons are provided in Section IV and the paper is concluded in Section V.

\section{sparse-smooth decomposition for image segmentation}

Sparsity-based signal decomposition has drawn a lot of attention in recent years because of its tremendous success in achieving the state-of-the-art results in many areas.
To decompose a signal into its components using sparse representation, usually two or more suitable dictionaries are found which are suitable for sparse representation of only one signal component, but not the others. 
As an example, to decompose a signal into two components as $x= x_1+x_2$, we need to find two over-complete dictionaries $T_1$ and $T_2$ which are suitable to sparsify $x_1$ and $x_2$ respectively. Then $x$ can be written as $x=T_1 \alpha_1 + T_2 \alpha_2$ by solving the following optimization problem:
\begin{gather*}
\argmin_{\alpha_1,\alpha_2} \|\alpha_1 \|_0 + \|\alpha_2 \|_0 \\
\text{subject to}  \ \ \ \| x-T_1 \alpha_1 - T_2 \alpha_2  \|_2 \leq \epsilon
\end{gather*}
After solving this optimization problem, $T_1 \alpha_1$ will correspond to the first component and $T_2 \alpha_2$ will correspond to the second component. The same technique can also be used for decomposition of a signal into more than two components.
In \cite{Starck}, Starck used a similar idea for image decomposition into texture and natural part. They also used total variation penalization to enforce smoothness to decomposition. 

In this paper, a segmentation algorithm is designed based on decomposition of the image into a smooth and a sparse component, which we call smooth-sparse decomposition.
It is designed based on the observation that the background part of the images mostly consist of smooth regions. Therefore they should be well-represented with a low order smooth model. By well-representation, we mean that the error of smooth approximation should be less than a desired threshold.

First the image is divided into non-overlapping blocks. Then each image block of size N$\times$N, $F(x,y)$, is represented with a smooth model $\hat{F}(x,y;\alpha_1,...,\alpha_K)$, where $x$ and $y$ denote the horizontal and vertical axes and $\alpha_1,...,\alpha_K$ are the parameters of smooth model.
Here $\hat{F}(x,y;\alpha_1,...,\alpha_K)$ is proposed to be chosen as a linear combination of $K$ smooth basis functions $\sum_{k=1}^K \alpha_k P_k(x,y)$, where $P_k(x,y)$ denotes a 2D smooth basis function. Here we use a set of low frequency two-dimensional DCT basis functions, since they are shown to be very suitable for natural image representation \cite{DCT} which is usually smooth. The 2-D DCT function is defined as:
\begin{equation*}
P_{u,v}(x,y)= \beta_u \beta_v cos((2x+1)\pi u/2N) cos((2y+1)\pi v/2N) 
\end{equation*}
where $u$ and $v$ denote the frequency of the basis. All basis functions are ordered in a zig-zag fashion in the (u,v) plane, and the first K basis functions are chosen.

Therefore the smooth approximation of an image block will be $\hat{F}(x,y)=\sum_{k=1}^K \alpha_k P_k(x,y)$.
As mentioned earlier, some parts of images cannot be represented with this smooth model, such as texts, lines and other high-frequency components which are overlaid on the smooth regions. To have a more general model which can represent both smooth and non-smooth parts of the image, we propose to add another component $S$ to the model as: 
\begin{equation}
\hat{F}(x,y)=\sum_{k=1}^K \alpha_k P_k(x,y)+S(x,y)
\end{equation}
where $\sum_{i=1}^K \alpha_i P_i(x,y)$ and $S(x,y)$ correspond to the smooth background region and foreground pixels respectively.  The $S$ component is usually sparse and contains the sharp edges and texts. So after performing this decomposition, those pixels which can be approximated as $\sum_{i=1}^K \alpha_i P_i(x,y)$ will be considered as background and the rest as foreground.

To have a more compact notation, we can look at the 1D version of this problem by converting the 2D blocks of size $N \times N$ into a vector of length $N^2$, denoted by $f$, and denoting $\sum_{k=1}^K \alpha_k P_k(x,y)$ as $ P\alpha$ where $P$ is a matrix of size $N^2\times K$ in which the k-th column corresponds to the vectorized version of $P_k(x,y)$ and $\alpha=[\alpha_1,...,\alpha_K]^\text{T}$. The 1D version of $S(x,y)$ is denoted by $s$.

Now to decompose the image, we need to derive the $\alpha_i$'s and $S$. To do so, we need to have some prior knowledge about these components. Here we proposed to solve this decomposition problem by maximizing the number of background pixels or equivalently minimizing the number of foreground pixels such that we do not try to model high frequency component in the image. This method is very desirable in image and video compression applications, because the background component can be easily represented using a set of low-order DCT bases. So the more background pixels we have, the smaller bit-rate we usually need. But we have to be careful not to represent the text and graphics by smooth model. To achieve this, we can have a regularization term which penalizes using too many basis.
Therefore we can solve the following optimization problem:
\begin{equation}
\begin{aligned}
& \underset{s, \alpha}{\text{minimize}}
& & \| s \|_0+ q \| \alpha \|_0  \\
& \text{subject to}
& & \| f-P \alpha-s  \|_2 \leq \epsilon
\end{aligned}
\end{equation}
where $q$ is a the scalar which controls the trade-off between these two terms.
Since L0 minimization is not a convex problem, we use L1 minimization as an approximation of L0 minimization which results in the following basis pursuit problem:
\begin{equation}
\begin{aligned}
& \underset{s, \alpha}{\text{minimize}}
& & \| s \|_1+q \| \alpha \|_1 \\
& \text{subject to}
& & \| f-P \alpha-s  \|_2 \leq \epsilon
\end{aligned}
\end{equation}
To further simplify this optimization, we can push $q$ in $P$ and minimize $\| s \|_1+\| \alpha \|_1$ instead of $\| s \|_1+q \| \alpha \|_1$. 
Therefore we can solve the following problem which is equivalent to (2):
\begin{equation}
\begin{aligned}
& \underset{s, \alpha}{\text{minimize}}
& & \| s \|_1+  \|\alpha \|_1  \\
& \text{subject to}
& & \| f-P^{'} \alpha-s  \|_2 \leq \epsilon
\end{aligned}
\end{equation}
where $P^{'}=\frac{1}{q} P$. Now this problem can be written in a more compact way by defining the new variable $y$ as the concatenated version of $s$ and $\alpha$, as $y^T=[ S \ \  \alpha]$ and defining $G_{N^2 \times (N^2+K)}$ as: $G= \big(  I_{N^2\times N^2} | P^{'}_{N^2\times K}  \big)$.
\begin{equation}
\begin{aligned}
& \underset{y}{\text{minimize}}
& & \| y\|_1  \\
& \text{subject to}
& & \| f-Gy \|_2 \leq \epsilon
\end{aligned}
\end{equation}
After solving this optimization problem, the first $N^2$ elements of $y$ will correspond to the $S$ component in the initial model and the rest of $y$ components correspond to $\alpha_i$'s. We can also look at the unconstrained version of this problem which is usually easier to solve by choosing a proper $\lambda$ as:

\begin{equation}
\begin{aligned}
& \underset{y}{\text{minimize}}
 \ \frac{1}{2} \| f-Gy  \|_2^2+ \lambda  \| y \|_1
\end{aligned}
\end{equation}
Solving the unconstrained problem is usually simpler than the constrained version, and in the remaining part of the paper, we focus on solving (5).
As it can be seen, this is the standard basis pursuit denoising problem and can be solved with iterative algorithms such as ISTA \cite{ISTA}, FISTA \cite{FISTA} and ADMM \cite{ADMM}. In this work, we have used ADMM algorithm to solve this problem.

\subsection{ADMM Formulation For LASSO}
ADMM (Alternating Direction Method of Multipliers) is a popular algorithm which combines superior convergence properties of method of multiplier and decomposability of dual ascent.
In the ADMM form, the optimization problem in (5) can be formulated as:
\begin{equation}
\begin{aligned}
& \underset{y,z}{\text{minimize}}
 \ \ \frac{1}{2} \| f-Gy  \|_2^2+ \lambda  \| z \|_1 \\
 & \text{subject to} \ \ y=z
\end{aligned}
\end{equation}
Then the following updates can be used for each iteration in ADMM \cite{ADMM}:
\begin{flalign*}
& y^{k+1}= (G^TG+\rho I)^{-1}(G^T f+\rho (z^k-u^k)) \\ 
& z^{k+1}= S_{\lambda/{\rho}}(y^{k+1}+u^k) \\
& u^{k+1}= u^k+y^{k+1}-z^{k+1}
\end{flalign*}
where $\rho$ is the augmented Lagrangian parameter and $S_{\lambda/{\rho}}$ denotes the soft-thresholding operator applied elementwise and is defined as:
\begin{gather*}
S_{\lambda/{\rho}}(x)= \text{sign}(x) \text{max}(|x|-\lambda/\rho,0)
\end{gather*}
We have to mention that the decomposition problem in (1) can also be done using robust regression which is studied in \cite{r_reg1} and \cite{r_reg2}.

\section{overall segmentation algorithm}
Some parts of images might be very simple to segment. Therefore we may not need to use sparse decomposition for segmentation of the entire image. Consequently we have some initial steps which first check if a block can be segmented using simpler approaches. These simpler approaches take care of two categories of blocks: entirely flat block in which all pixels have the same value and are common in screen content images and smoothly varying background without foreground in which the intensity of all pixels can be modeled well by the smooth function.
Entirely flat blocks can be declared as background or foreground based on their neighboring blocks' background color. For these blocks, if we could find at least one neighbor block with a background color close enough to the current block's color (with a difference less than $\epsilon_2$), it would be segmented as background.
For smoothly varying background, we try to fit $K$ DCT basis to all pixels using least square fitting and if the intensity of all pixels can be predicted with distortion less than $\epsilon_3$, that block would be segmented entirely as background.
We will use the sparse decomposition method only if a block does not satisfy these two conditions. 

The segmentation algorithm is summarized below:

\begin{enumerate}
\item If all pixels in the block have the same color intensity (i.e. it is entirely flat block), declare the entire block as background or foreground as explained above. If not, go to the next step;
\item Perform least square fitting using the luminance values of all pixels. If all pixels can be predicted with an absolute error less than $\epsilon_3$, declare the entire block as background. If not, go to the next step;
\item Use sparse-smooth decomposition to fit a model to the luminance values of image block and find the absolute fitting error of all pixels using the smooth model. Each pixel with a distortion less than a threshold $\epsilon_1$ will be considered as background, otherwise as foreground. 
\end{enumerate}

\section{Results}

We have tested our algorithm on several test images selected from HEVC standard test sequences for screen content coding.
We have generated a dataset consisting of 332 image blocks of size $64\times 64$. The ground truth foreground maps for these images are extracted manually. This dataset is publicly available and can be downloaded from \cite{dataset}.

Before showing the results, let us discuss about the parameters used in our algorithm. 
The block size and the number of DCT bases are chosen to be $N$=64 and $K$=10 respectively. The weight parameter $q$ is chosen to be $q=\frac{1}{100}$.
For the segmentation steps, the parameters are chosen as $\epsilon_1$, $\epsilon_2=10$ and $\epsilon_3=3$, which have been found to perform well on a training set. For the ADMM algorithm, the implementation by Stephen Boyd \cite{boyd} is used. The number of iterations is chosen to be 100 and the parameter $\rho$ is set to 1 as the default value.

We have compared the segmentation result of the proposed algorithm with hierarchical k-means clustering in \cite{djvu} and shape primitive extraction and coding in \cite{SPEC}. 
To provide a numerical comparison between the proposed scheme and the previous approaches, we have calculated the average precision and recall achieved by different segmentation algorithms over this dataset. The average precision and recall by different algorithms are given in Table 1. As it can be seen, the proposed scheme achieves a much higher precision and recall than other algorithms.

\begin{table} [h]
\centering
  \caption{Comparison of performance of different algorithms}
  \centering
\begin{tabular}{|m{3.2cm}|m{2cm}|m{2cm}|}
\hline
 \ \ Different Algorithms  &  \ \ \ \ \ Precision & \ \ \ \ \ \ Recall\\
\hline
\ \ SPEC \cite{SPEC} & \ \ \ \ \ \ 0.50 & \ \ \ \ \ \  0.64 \\
\hline
\ \ Hierarchical Clustering \cite{djvu} & \ \ \ \ \ \ 0.64 & \ \ \ \ \ \ 0.69 \\
\hline
\ \ Proposed Scheme & \ \ \ \ \ \ 0.64 & \ \ \ \ \ \ 0.95 \\
\hline
\end{tabular}
\label{TblComp}
\end{table}

\begin{figure*}[ht]
        \centering
        \vspace{-0.5cm}
        \begin{subfigure}[b]{0.18\textwidth}
                \includegraphics[width=\textwidth]{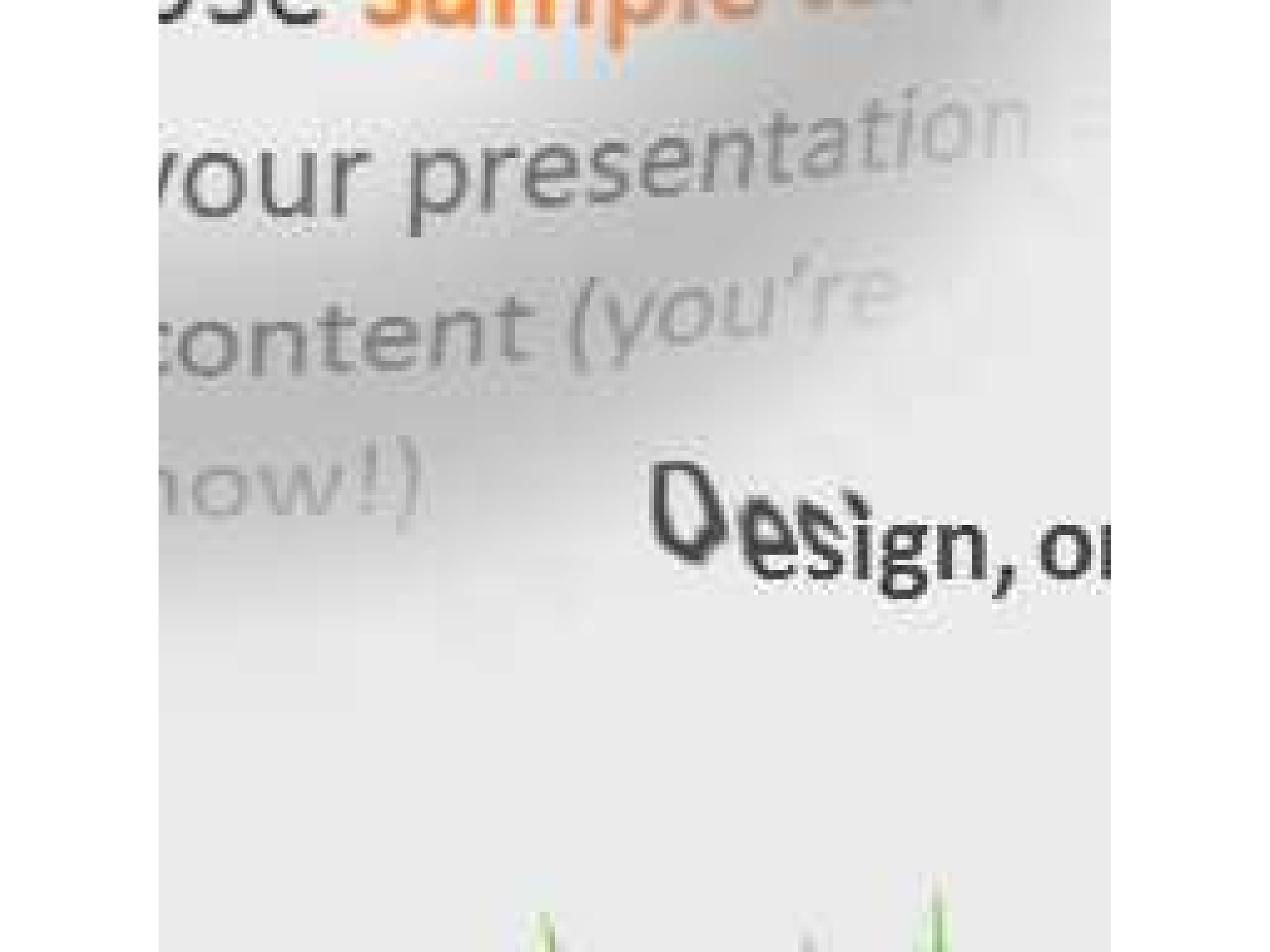}
                                \vspace{-0.5cm}
          \hspace{-2.5cm}    
        \end{subfigure}%
        ~ 
        \begin{subfigure}[b]{0.18\textwidth}
                \includegraphics[width=\textwidth]{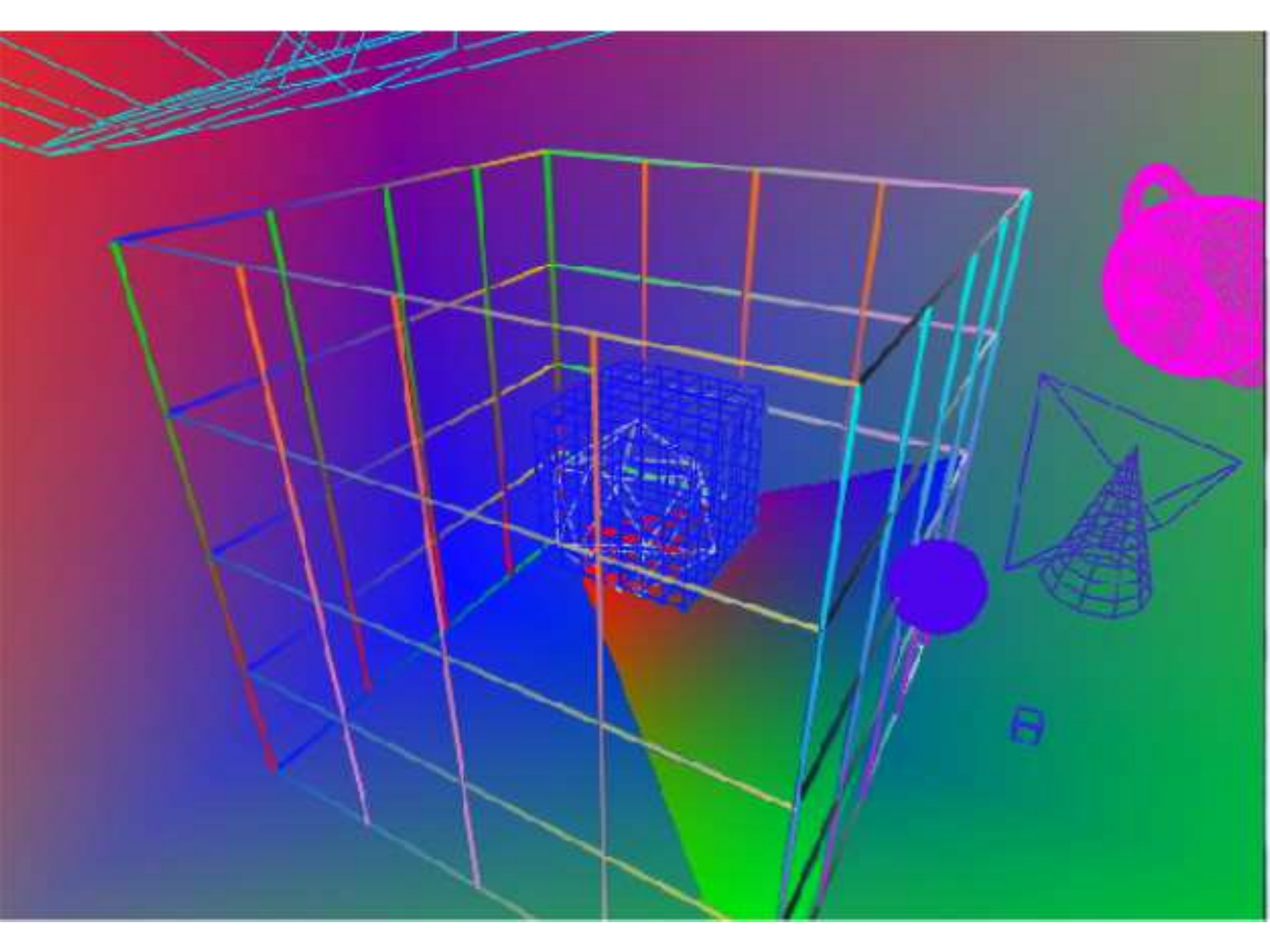}
                \vspace{-0.5cm}
            \hspace{-3cm} 
        \end{subfigure}%
        ~ 
        \begin{subfigure}[b]{0.18\textwidth}
                \includegraphics[width=\textwidth]{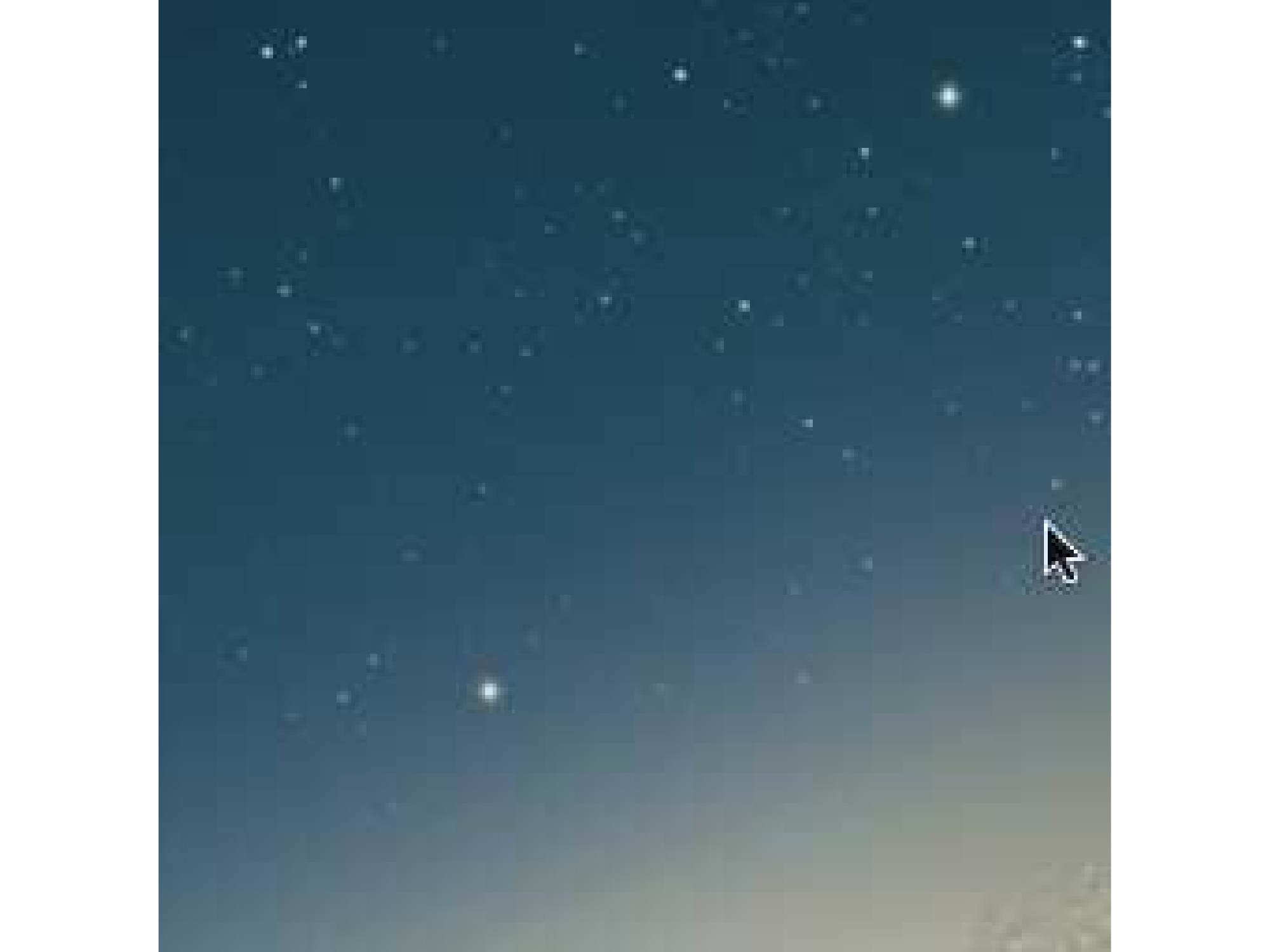}
                \vspace{-0.45cm}
            \hspace{-3cm} 
        \end{subfigure}%
        \begin{subfigure}[b]{0.18\textwidth}
			~ 
                \includegraphics[width=\textwidth]{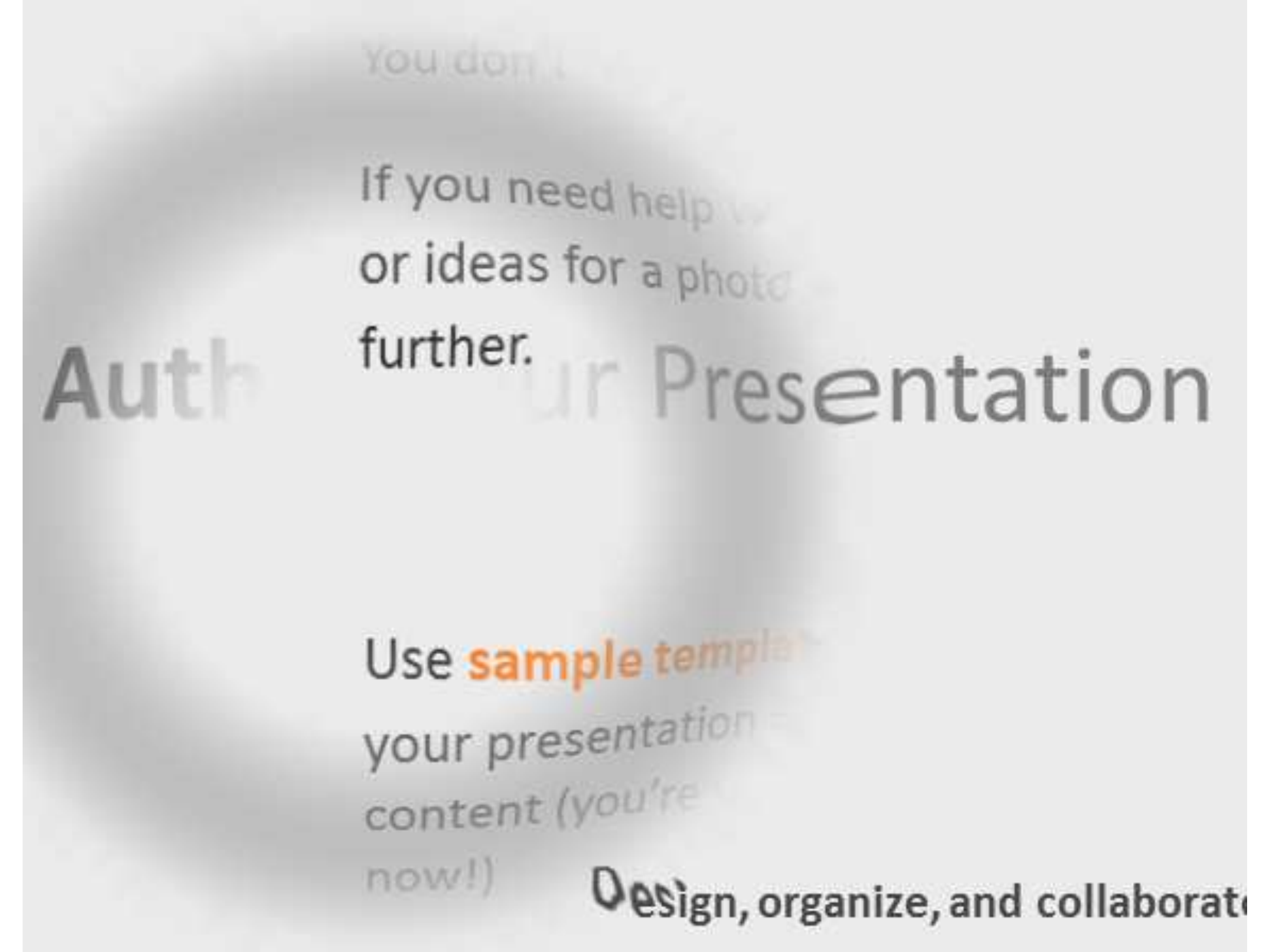}
                \vspace{-0.45cm}
            \hspace{-5cm} 
        \end{subfigure}%
        \begin{subfigure}[b]{0.18\textwidth}
                \includegraphics[width=\textwidth]{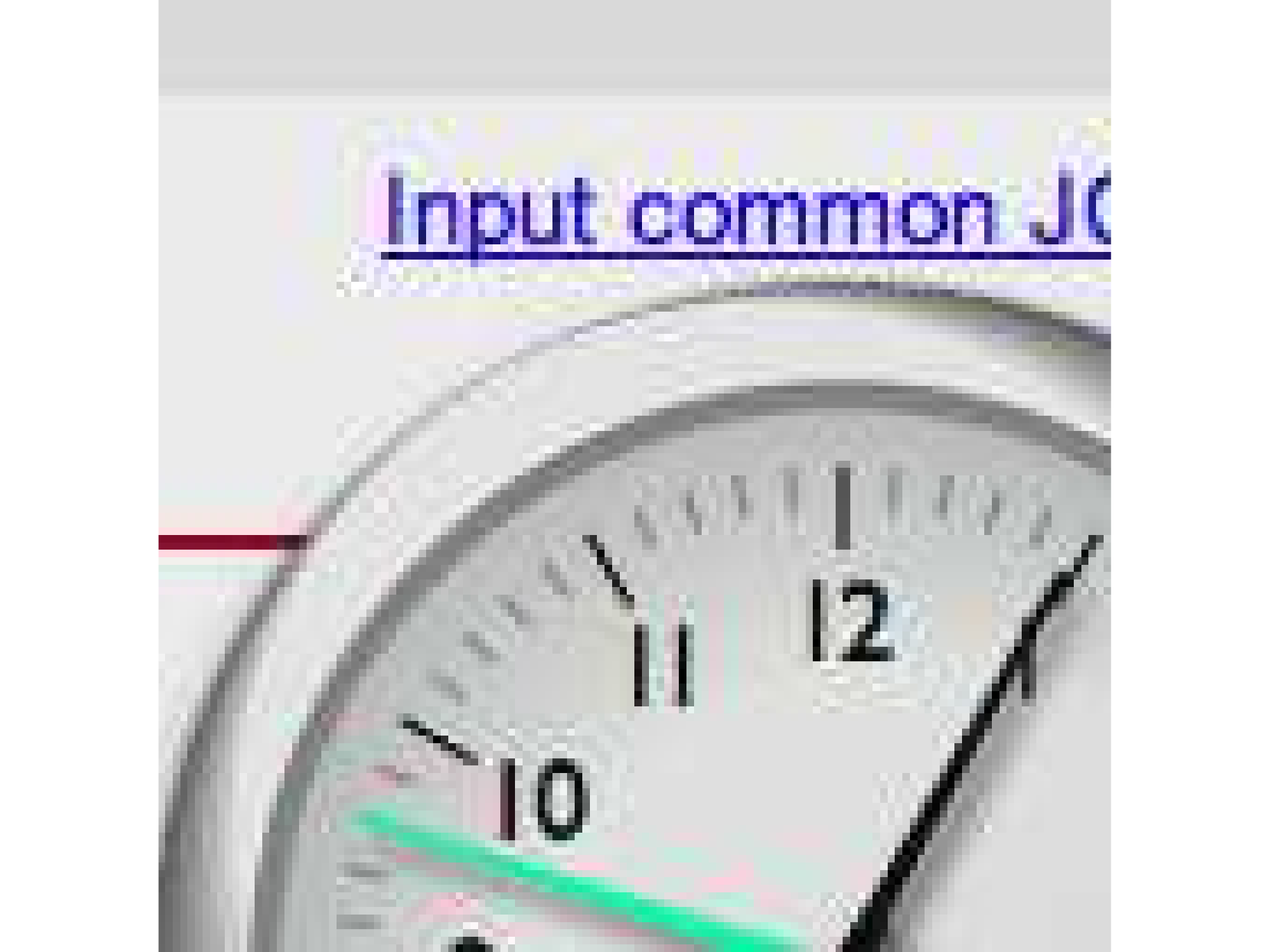}
                 \vspace{-0.45cm}
              \hspace{-4.8cm}
        \end{subfigure}
         \\[1ex]
        \begin{subfigure}[b]{0.18\textwidth}
                \includegraphics[width=\textwidth]{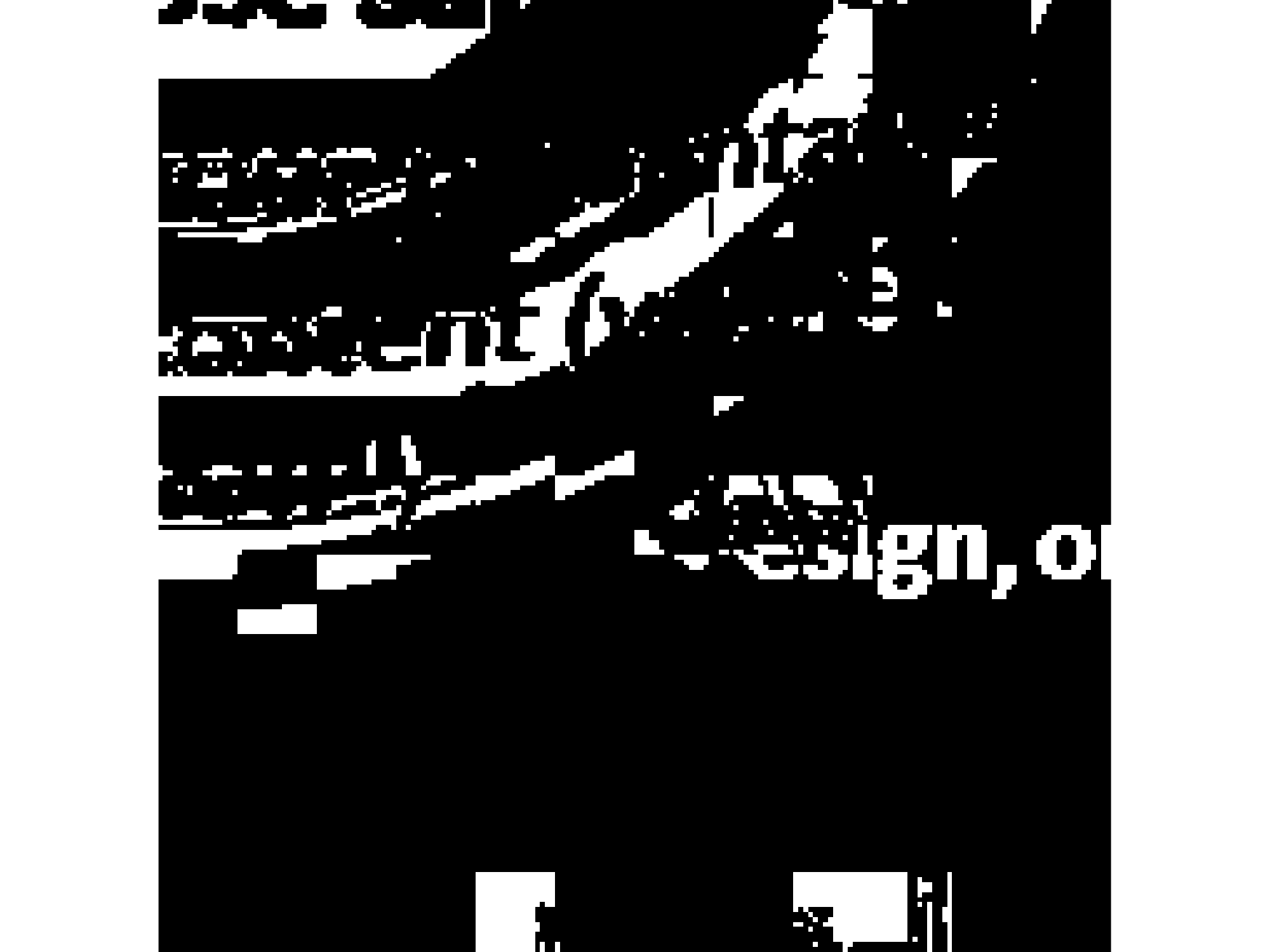}
                                \vspace{-0.5cm}
          \hspace{-2.5cm}    
        \end{subfigure}%
        ~ 
        \begin{subfigure}[b]{0.18\textwidth}
                \includegraphics[width=\textwidth]{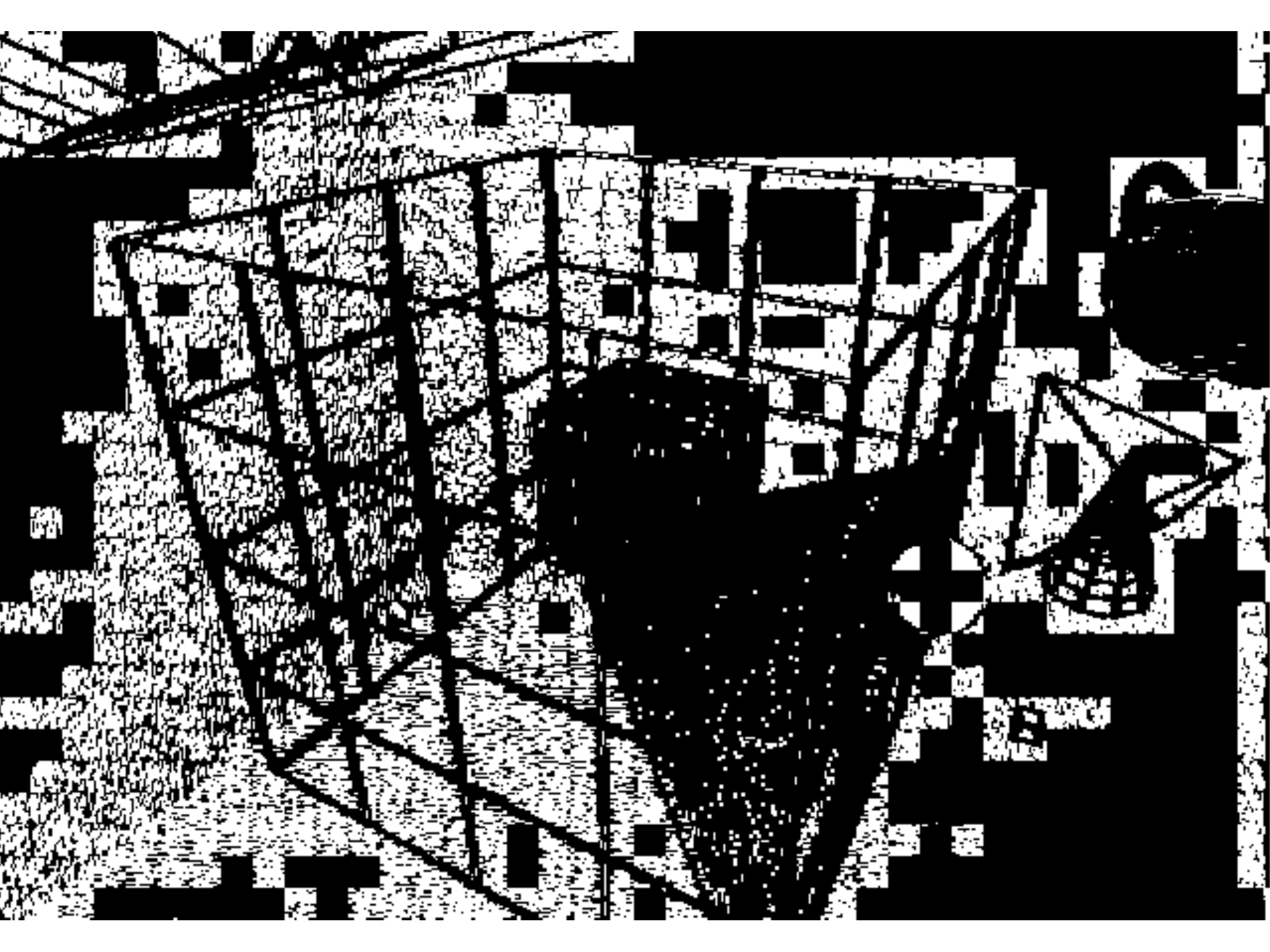}
                \vspace{-0.5cm}
            \hspace{-3cm} 
        \end{subfigure}%
        ~ 
        \begin{subfigure}[b]{0.18\textwidth}
                \includegraphics[width=\textwidth]{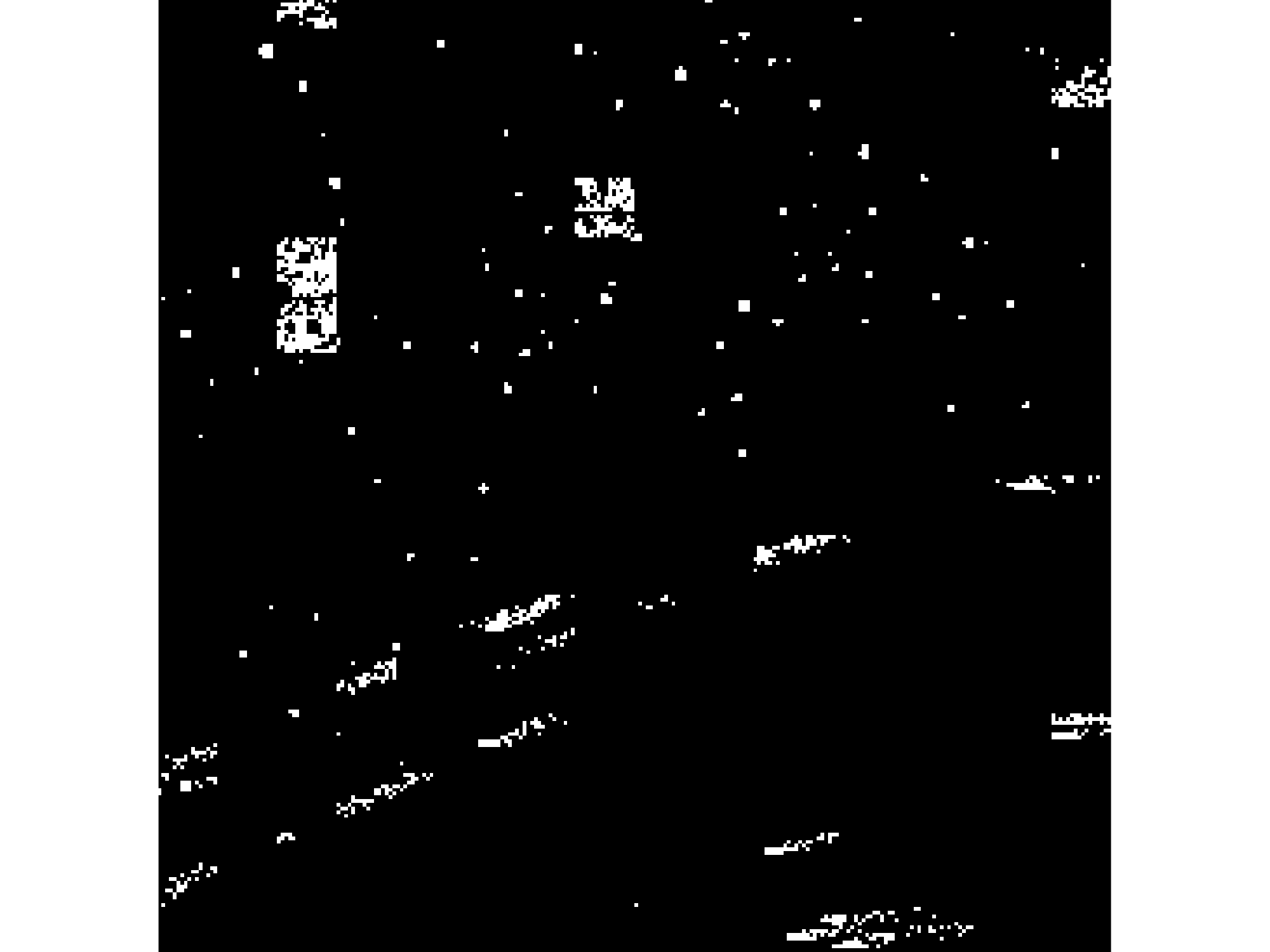}
                \vspace{-0.45cm}
            \hspace{-3cm} 
        \end{subfigure}%
        \begin{subfigure}[b]{0.18\textwidth}
			~ 
                \includegraphics[width=\textwidth]{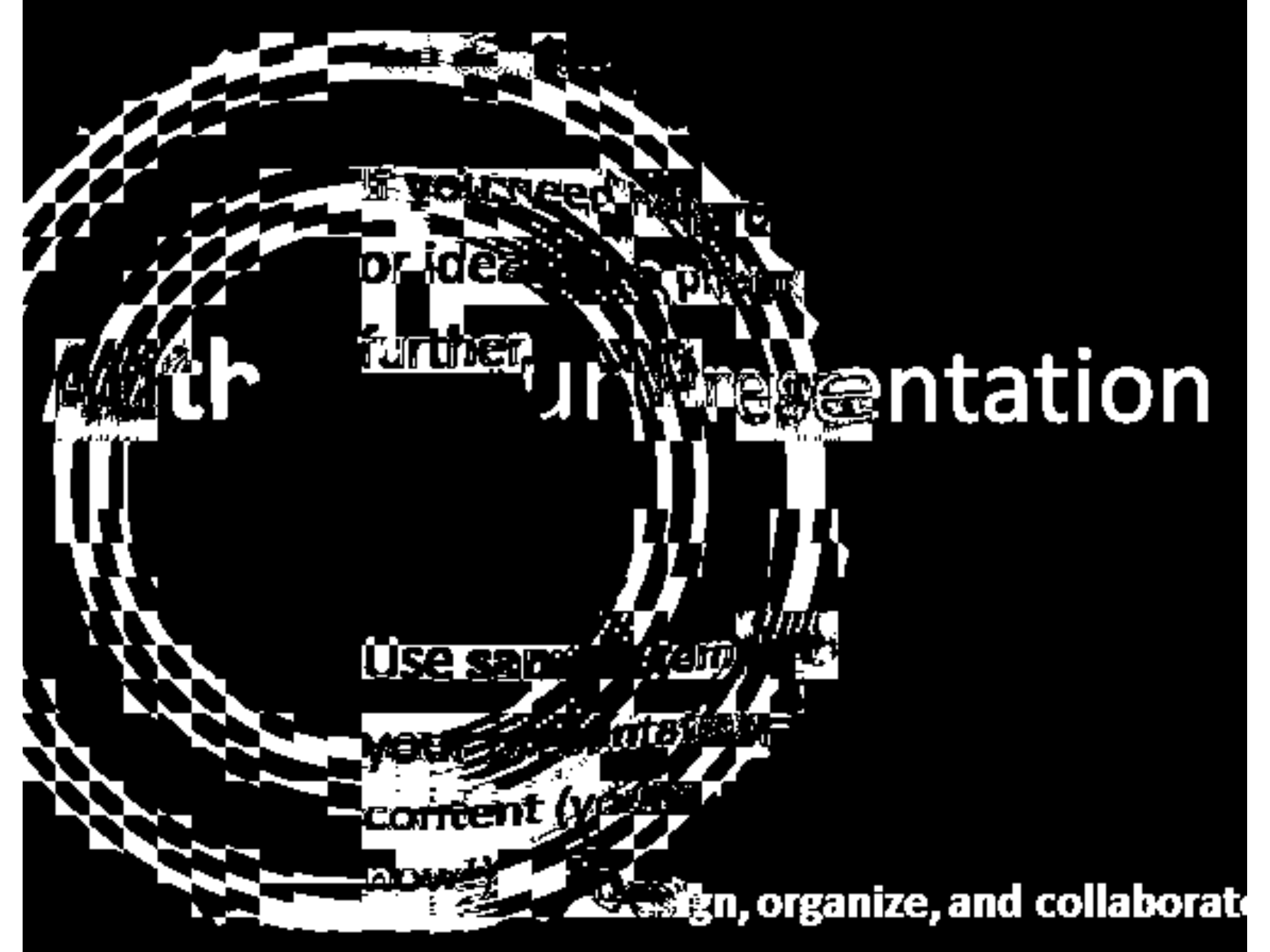}
                \vspace{-0.45cm}
            \hspace{-3cm} 
        \end{subfigure}%
        \begin{subfigure}[b]{0.18\textwidth}
                \includegraphics[width=\textwidth]{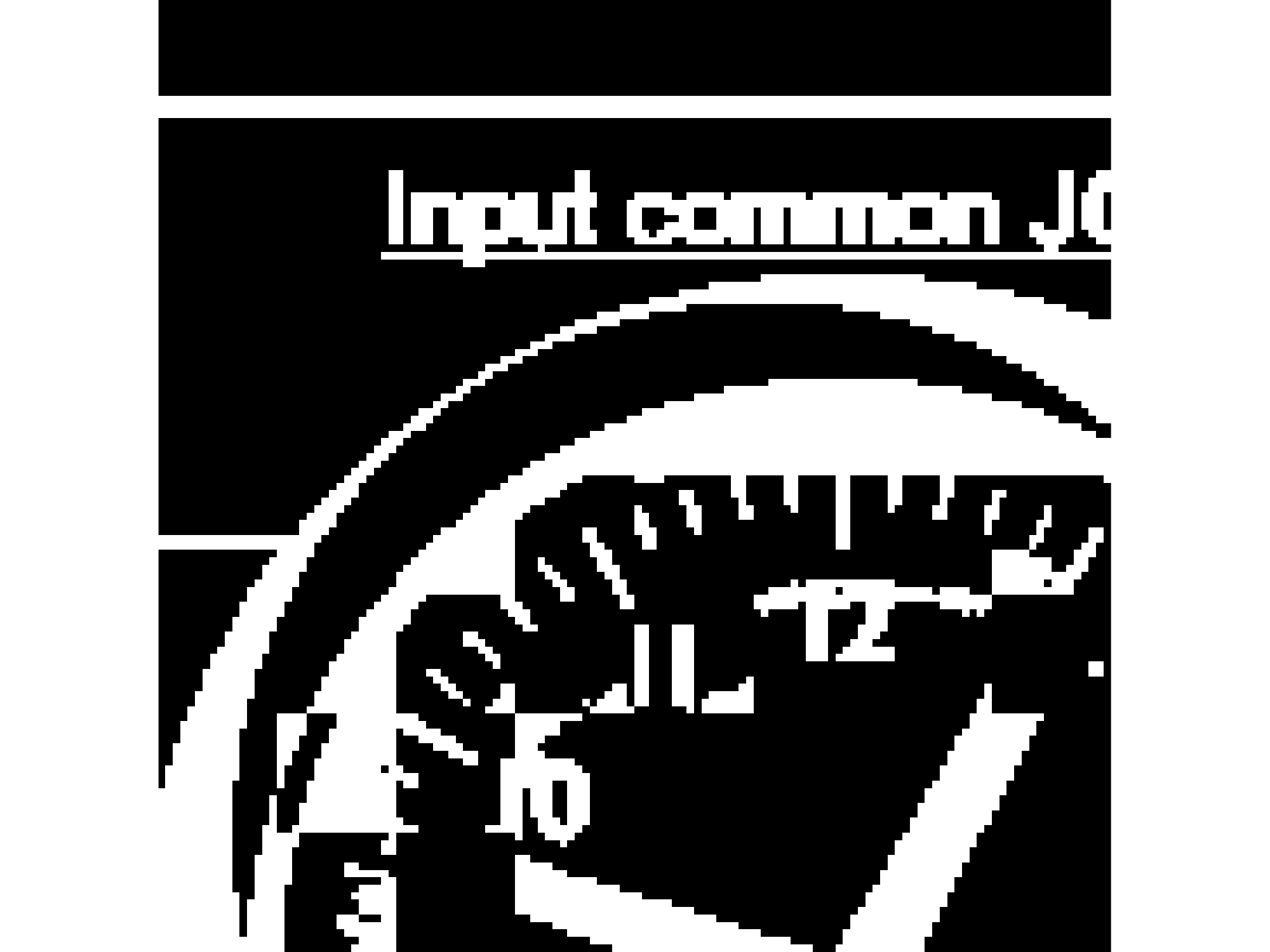}
                 \vspace{-0.45cm}
              \hspace{-4.8cm}
        \end{subfigure}\\[1ex]
        \begin{subfigure}[b]{0.18\textwidth}
                \includegraphics[width=\textwidth]{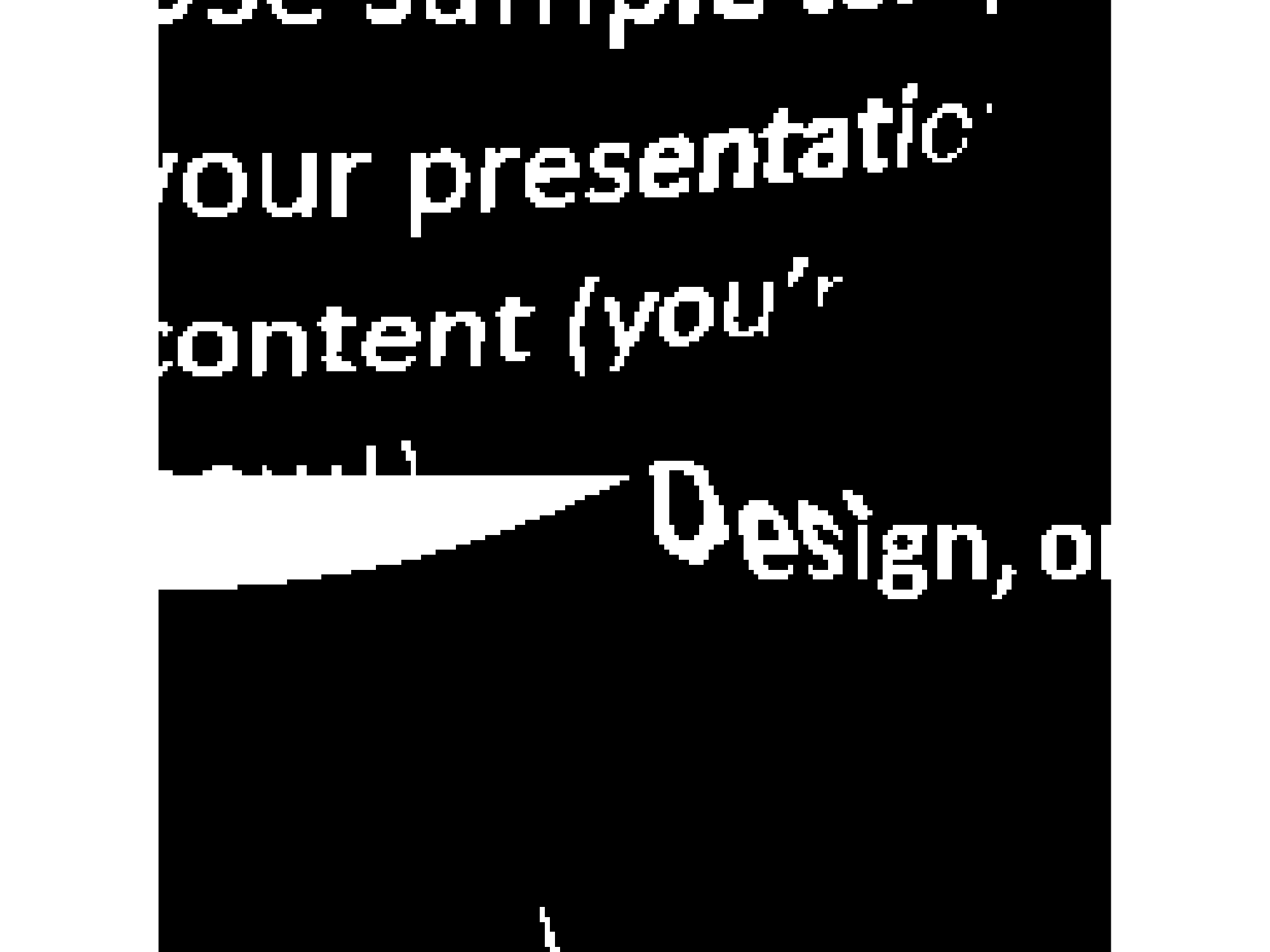}
                                \vspace{-0.5cm}
          \hspace{-2.5cm}    
        \end{subfigure}%
        ~ 
        \begin{subfigure}[b]{0.18\textwidth}
                \includegraphics[width=\textwidth]{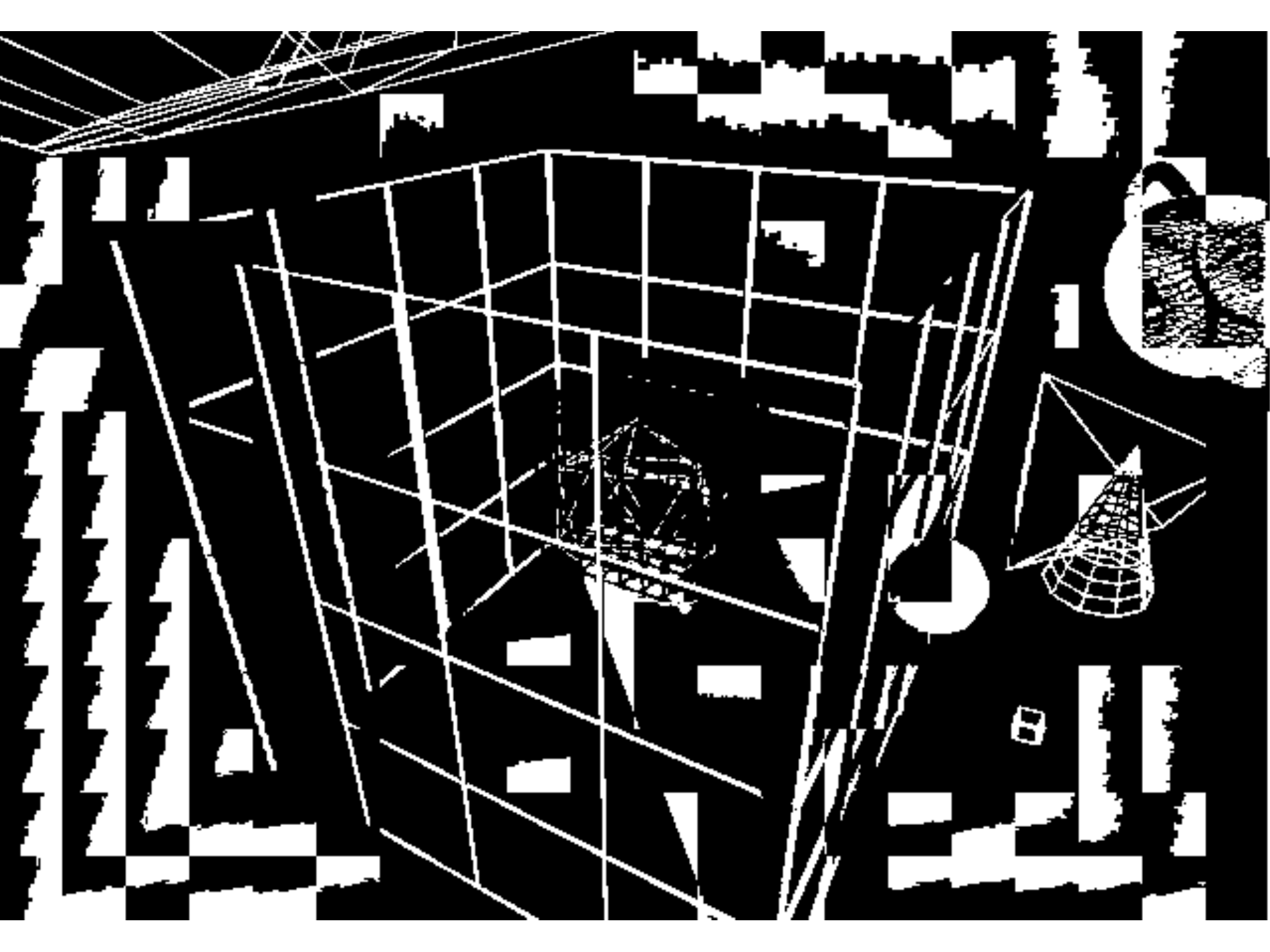}
                \vspace{-0.5cm}
            \hspace{-3cm} 
        \end{subfigure}%
        ~ 
        \begin{subfigure}[b]{0.18\textwidth}
                \includegraphics[width=\textwidth]{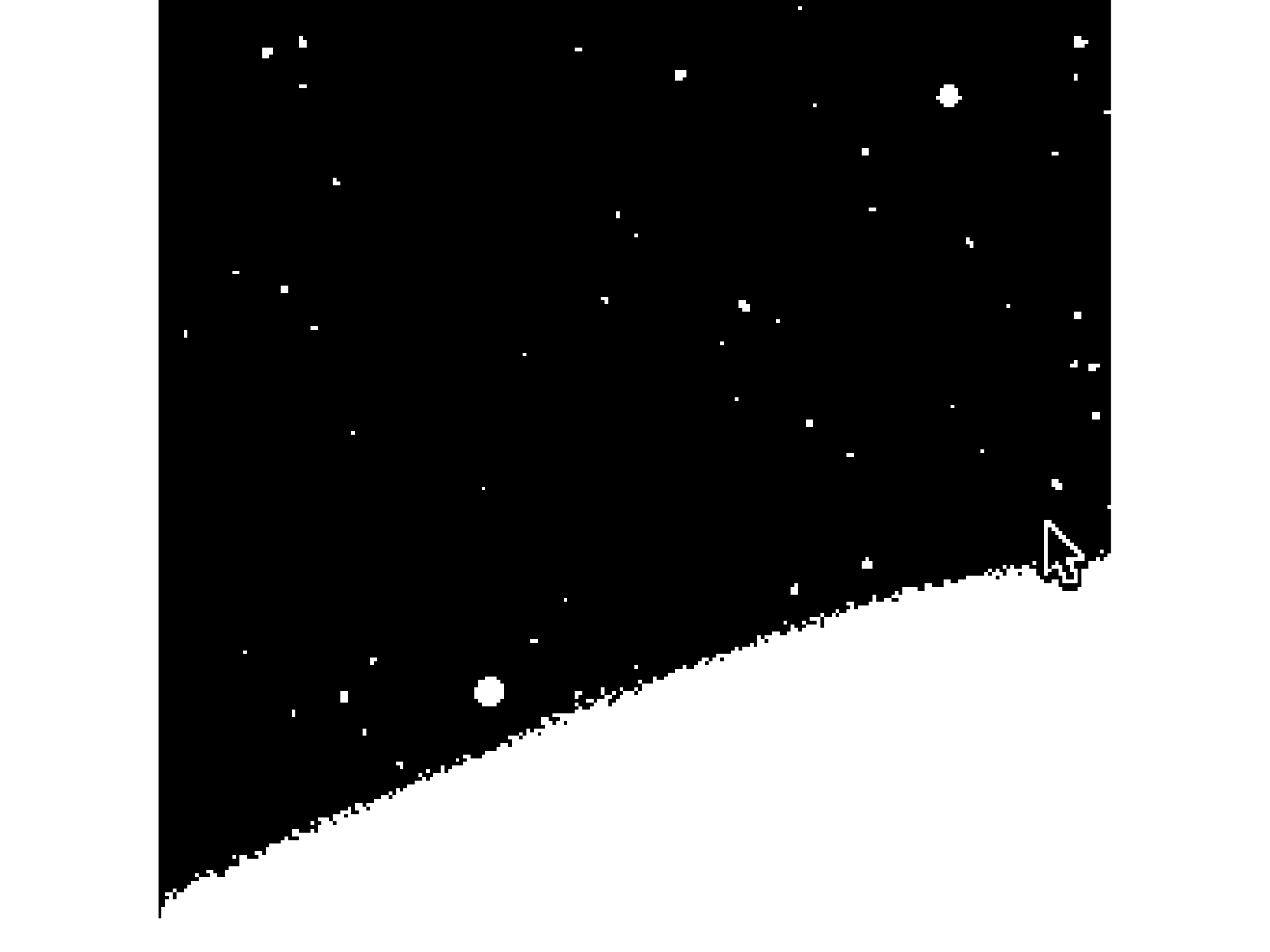}
                \vspace{-0.45cm}
            \hspace{-3cm} 
        \end{subfigure}%
        \begin{subfigure}[b]{0.18\textwidth}
			~ 
                \includegraphics[width=\textwidth]{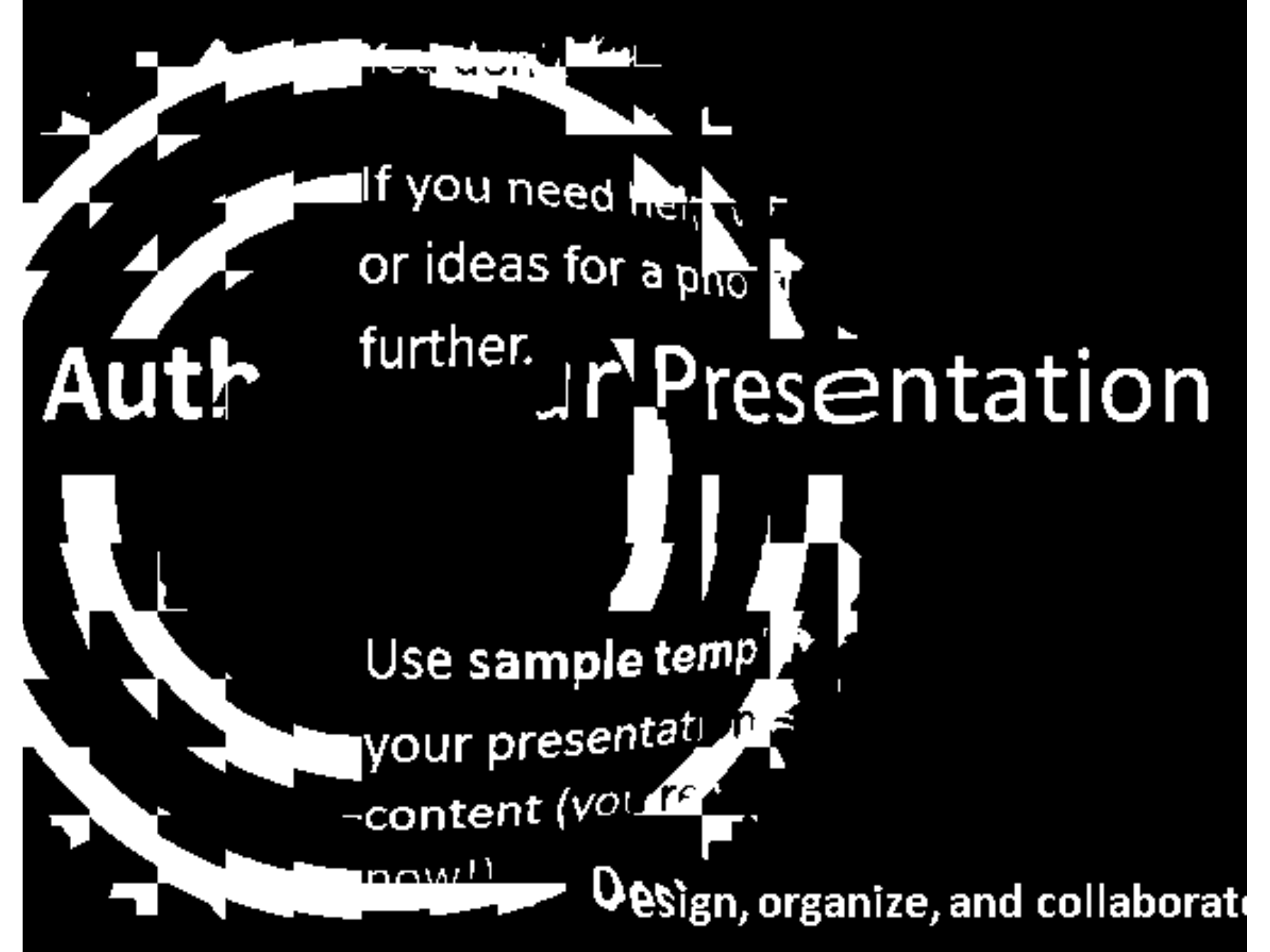}
                \vspace{-0.45cm}
            \hspace{-3cm} 
        \end{subfigure}%
        \begin{subfigure}[b]{0.18\textwidth}
                \includegraphics[width=\textwidth]{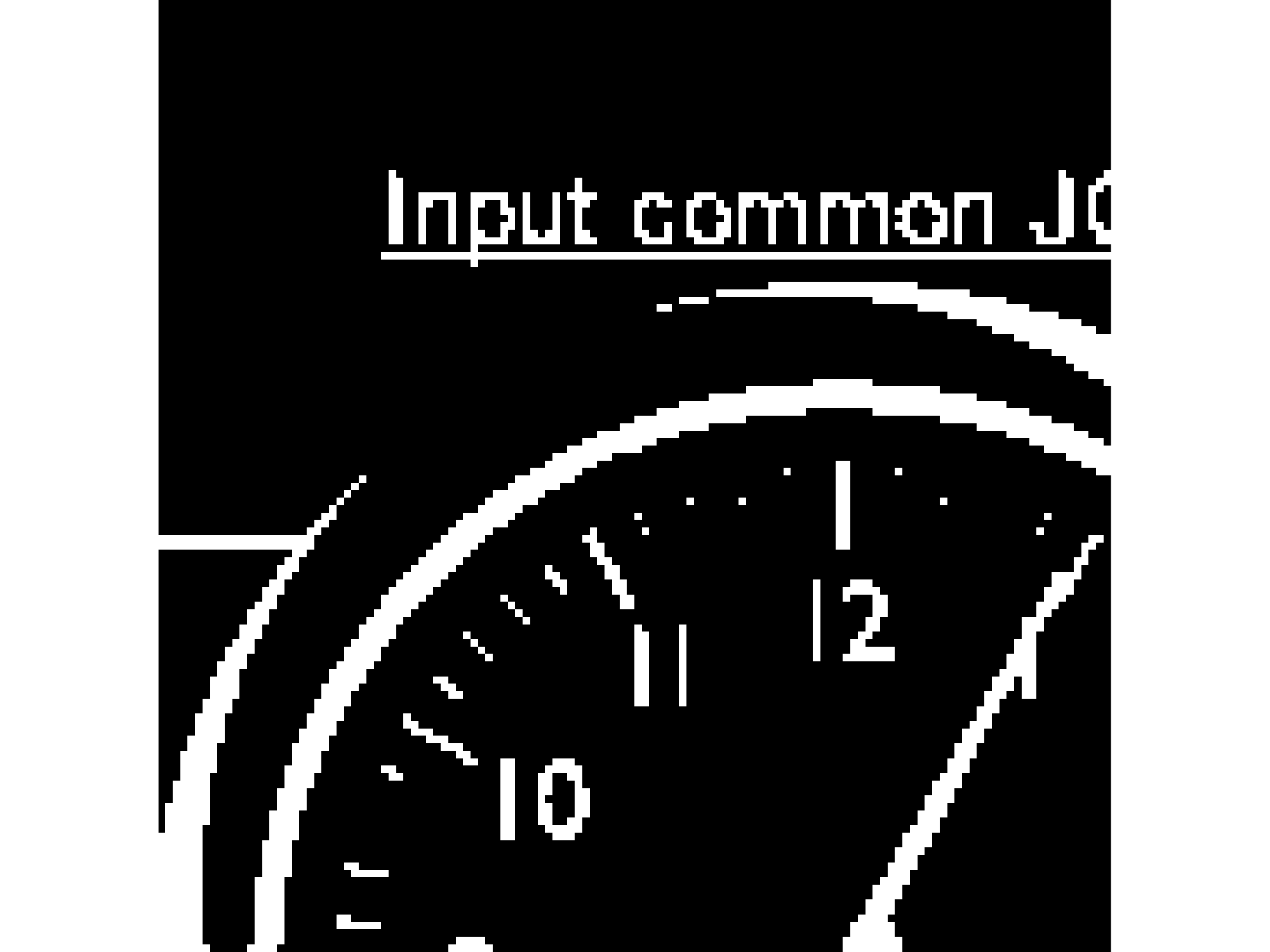}
                 \vspace{-0.45cm}
              \hspace{-4.8cm}
        \end{subfigure} \\[1ex]        
        \begin{subfigure}[b]{0.18\textwidth}
                \includegraphics[width=\textwidth]{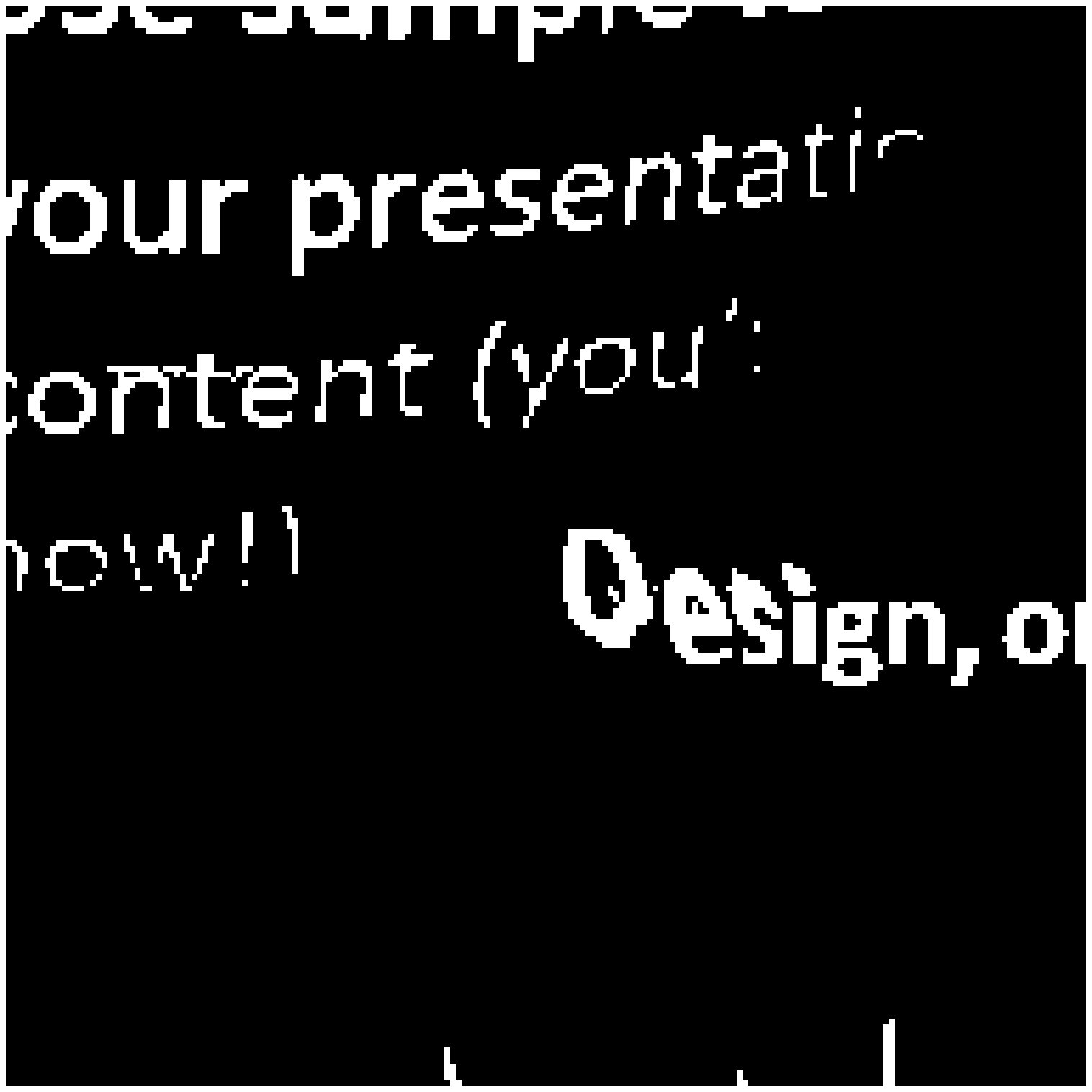}
                                \vspace{-0.5cm}
          \hspace{-2.5cm}    
        \end{subfigure}%
        ~ 
        \begin{subfigure}[b]{0.18\textwidth}
                \includegraphics[width=\textwidth]{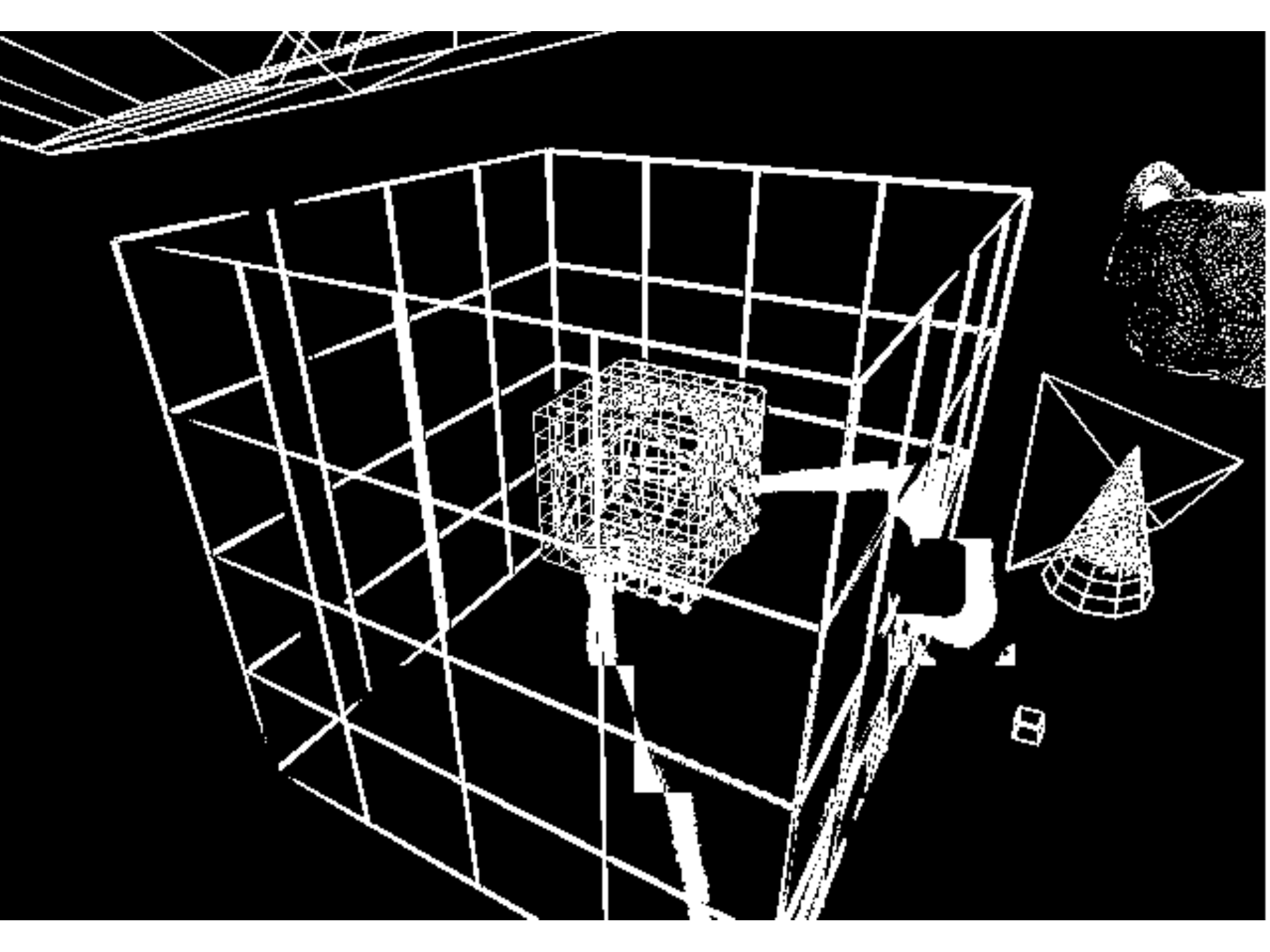}
                \vspace{-0.5cm}
            \hspace{-3cm} 
        \end{subfigure}%
        ~ 
        \begin{subfigure}[b]{0.18\textwidth}
                \includegraphics[width=\textwidth]{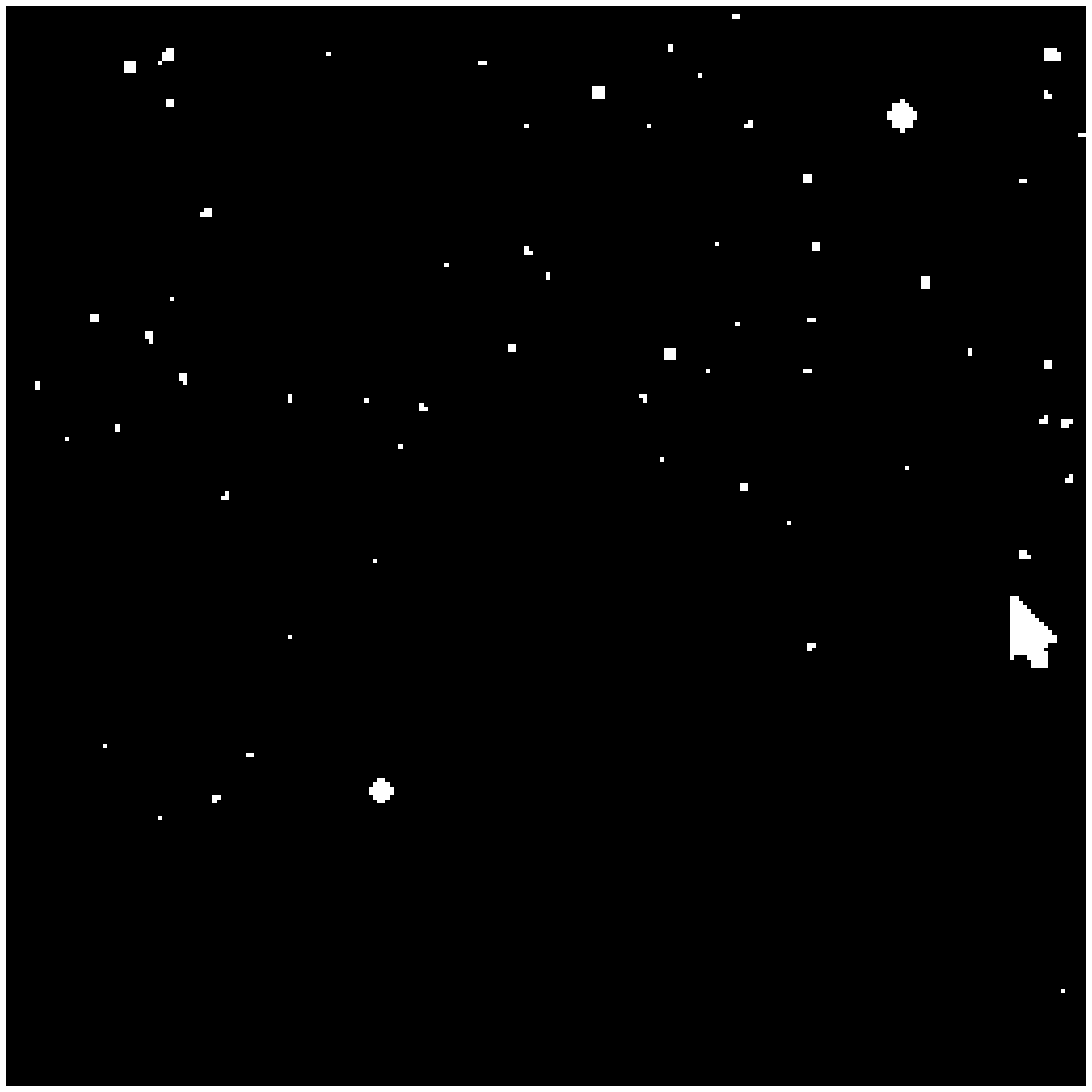}
                \vspace{-0.45cm}
            \hspace{-3cm} 
        \end{subfigure}%
        \begin{subfigure}[b]{0.18\textwidth}
			~ 
                \includegraphics[width=\textwidth]{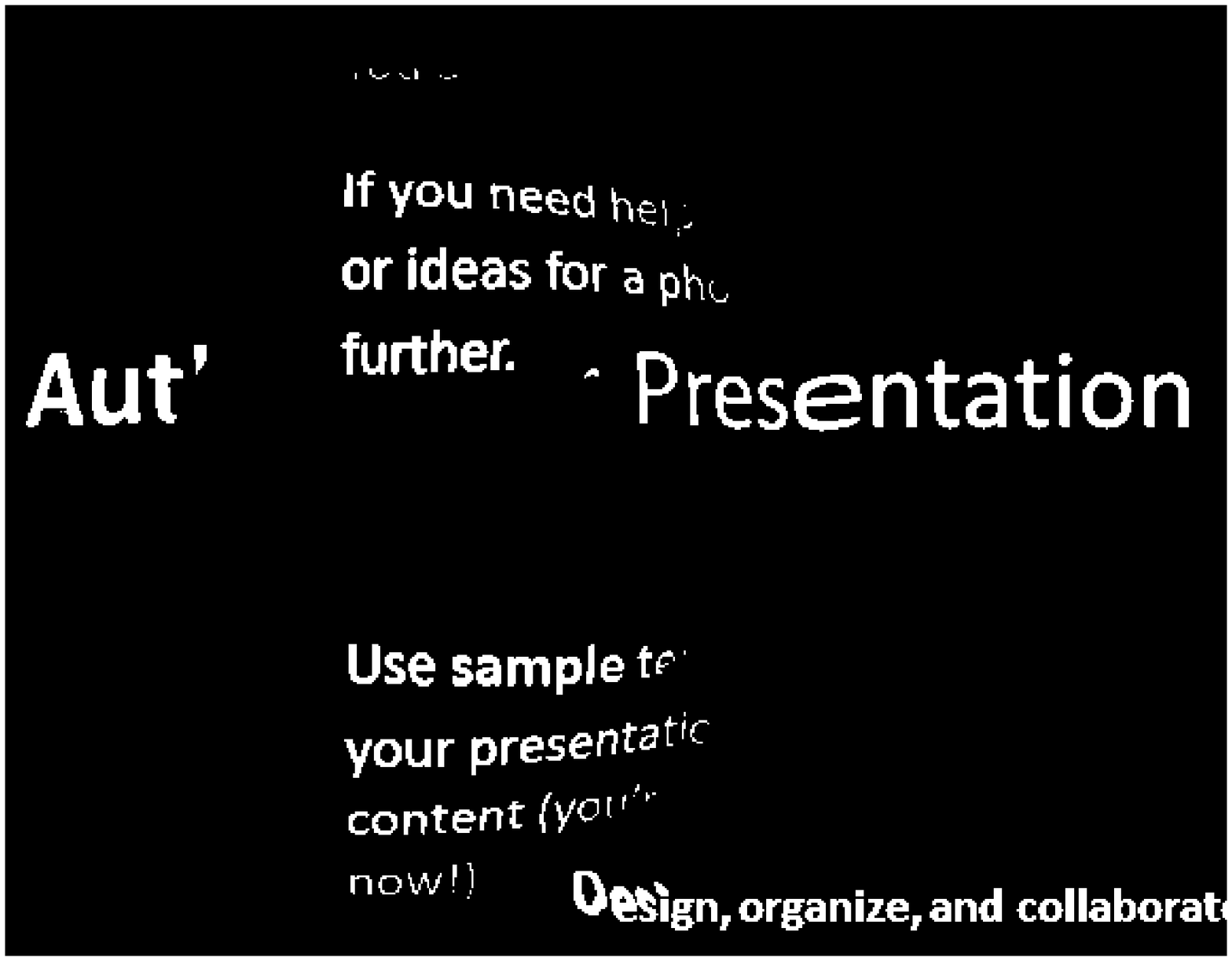}
                \vspace{-0.45cm}
            \hspace{-3cm} 
        \end{subfigure}%
        \begin{subfigure}[b]{0.18\textwidth}
                \includegraphics[width=\textwidth]{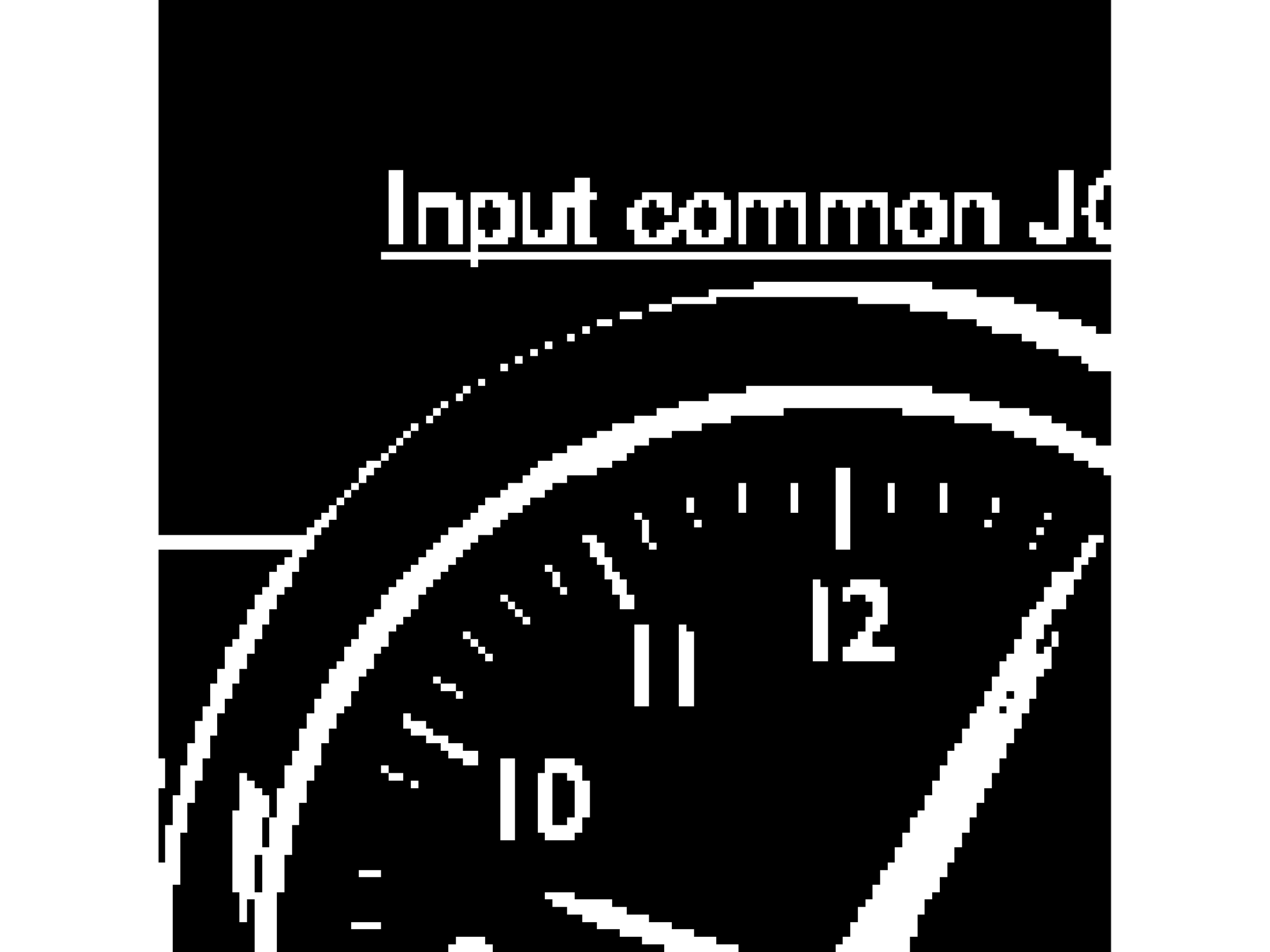}
                 \vspace{-0.45cm}
              \hspace{-4.8cm}
        \end{subfigure}
        \caption{Segmentation result for test images. The images in the first row denote the original images. The images in the second, third and forth rows denote the foreground map by SPEC, hierarchical clustering in DjVu and the proposed algorithm respectively.}
\end{figure*}

Figure 1 shows the segmentation result by different algorithms for 5 test images. As it can be seen, the proposed scheme outperforms both the hierarchical k-means clustering and SPEC algorithm in terms of visual quality of the segmentation.
We have mainly used challenging images in our dataset where the background and foreground have overlapping color ranges. For simpler cases where the background has a narrow color range that is quite different from the foreground, hierarchical k-means clustering algorithm will also perform well.

\section{Conclusion}
In this paper, an algorithm for separation of background and foreground is proposed, assuming that the background usually contains the smooth parts of the image and foreground contains text, lines and graphics. This algorithm decomposes the image into two components, smooth and sparse, which correspond to the background and foreground layers respectively.
A regularized L1 optimization problem is used to perform this decomposition. A pixel is considered background if it can be represented accurately by the smooth model; otherwise it will be considered as foreground. 
To speed up the algorithm, we developed some initial steps to segment the simple blocks.
The proposed scheme has been tested on a set of screen content images and is compared with two other algorithms, DjVu and SPEC, exhibiting superior performance for cases where foreground and background colors overlap.
Besides screen content image segmentation, this algorithm can be utilized in text extraction from images and medical image segmentation.

\end{document}